# Large-Scale Pre-training Enables Multimodal AI Differentiation of Radiation Necrosis from Brain Metastasis Progression on Routine MRI

Running Title: AI Differentiation of Radionecrosis vs. Recurrence

**Authors:**

Ahmed Gomaa, Annette Schwarz, Ludwig Singer, Arnd Dörfler, Matthias Stefan May, Pluvio Stephan, Ishita Sheth, Juliane Szkitsak, Katharina Breininger, Yixing Huang, Benjamin Frey, Oliver Schnell, Daniel Delev, Roland Coras, Daniel Höfler, Philipp Schubert, Jenny Stritzelberger, Sabine Semrau, Andreas Maier, Dieter H Heiland, Udo S. Gaipl, Andrea Wittig, Rainer Fietkau, Christoph Bert, Stefanie Corradini, and Florian Putz

**Affiliations:**

Department of Radiation Oncology, University Hospital Erlangen, Friedrich-Alexander-Universität Erlangen-Nürnberg, Erlangen, Germany (A.G, A.S, P.S, I.S, J.S., B.F, D.H, P.S., S.S., U.S.G, R.F, C.B, S.C, F.P)

Comprehensive Cancer Center Erlangen-EMN (CCC ER-EMN), Erlangen, Germany (A.G, A.S, P.S, I.S, J.S., B.F, D.H, P.S., S.S., U.S.G, R.F, C.B, S.C, F.P)

Bavarian Cancer Research Center (BZKF), Erlangen, 91052, Germany (A.G, A.S, P.S, I.S, J.S., B.F, A.D., O.S., D.D., R.C., D.H, P.S., M.S.M., L.S., J.S., S.S., D.H.H., U.S.G, R.F, C.B, S.C, F.P)

Center for Artificial Intelligence and Data Science, Universität Würzburg, Würzburg, 97074, Germany (K.B)

Health Science Center, Institute of Medical Technology, Peking University (Y.H)

Institute of Neuroradiology, University Hospital Erlangen, Friedrich-Alexander-Universität Erlangen-Nürnberg, Erlangen, Germany (A.D, L.S)

Department of Neurosurgery, University Hospital Erlangen, Friedrich-Alexander-Universität Erlangen-Nürnberg, Erlangen, Germany (O.S, D.D, R.C, D.H.H)

Department of Neurology, University Hospital Erlangen, Friedrich-Alexander-Universität Erlangen-Nürnberg, Erlangen, Germany (J.S)

Pattern Recognition Lab, Friedrich-Alexander- Universität Erlangen-Nürnberg, Erlangen, Germany (A.M)

Klinik und Poliklinik für Strahlentherapie und Radioonkologie, Universitätsklinikum Würzburg, Würzburg, Germany. (A.W.)

**Corresponding Author:**

PD Dr. Florian Putz
Department of Radiation Oncology
Universitätsklinikum Erlangen
Universitätsstraße 27
91054 Erlangen, Germany
Email: florian.putz@uk-erlangen.de
ORCID: 0000-0003-3966-2872

## Abstract

**Background**: Differentiating radiation necrosis (RN) from tumor progression after stereotactic radiosurgery (SRS) remains a critical challenge in brain metastases. While histopathology represents the gold standard, its invasiveness limits feasibility. Conventional supervised deep learning approaches are constrained by scarce biopsy-confirmed training data. Self-supervised learning (SSL) overcomes this by leveraging the growing availability of large-scale unlabeled brain metastases imaging datasets.

**Methods**: In a two-phase deep learning strategy inspired by the foundation model paradigm, a Vision Transformer (ViT) was pre-trained via SSL on 10,167 unlabeled multi-source T1CE MRI sub-volumes. The pre-trained ViT was then fine-tuned for RN classification using a two-channel input (T1CE MRI and segmentation masks) on the public MOLAB dataset (n=109) using 20% of datasets as same-center held-out test set. External validation was performed on a second-center test cohort (n=28).

**Results**: The self-supervised model achieved an AUC of 0.916 on the same-center test set and 0.764 on the second center test set, surpassing the fully supervised ViT (AUC 0.624/0.496; p=0.001/0.008) and radiomics (AUC 0.807/0.691; p=0.005/0.014). Multimodal integration further improved performance (AUC 0.947/0.821; p=0.073/0.001). Attention map visualizations enabled interpretability showing the model focused on clinically relevant lesion subregions.

**Conclusion**: Large-scale pre-training on increasingly available unlabeled brain metastases datasets substantially improves AI model performance. A two-phase multimodal deep learning strategy achieved high accuracy in differentiating radiation necrosis from tumor progression using only routine T1CE MRI and standard clinical data, providing an interpretable, clinically accessible solution that warrants further validation.

## Key Points

- Large-scale pre-training on unlabeled brain metastases MRI datasets overcomes training data scarcity
- A multimodal AI model accurately differentiated radiation necrosis from progression using T1CE MRI and clinical data alone
- Interpretability through transformer attention maps

## Importance of the study

This study addresses the critical challenge of distinguishing radiation necrosis from tumor progression in brain metastasis patients after radiosurgery. The current gold standard, histopathology, is often unfeasible due to its invasive nature. Alternative advanced imaging techniques are not universally available. Computational models applicable to standard imaging on the other hand lack generalizability because they require large, biopsy-confirmed datasets for training, which are scarce.

This research demonstrates a novel approach using self-supervised learning. This method leverages vast amounts of unlabeled, routine MRI scans to build a robust model. The resulting tool provides a non-invasive and interpretable way to solve this challenging task. Most importantly, it uses only standard T1-weighted MRI and clinical data. This makes the solution accessible for widespread clinical adoption and promising for guiding critical treatment decisions without invasive procedures.

# Introduction

Brain metastases (BMs) are the most common intracranial tumors in adults, occurring in roughly 20% to 40% of all cancer patients.[1,2] Advances in systemic therapies have improved patient survival, in turn increasing the incidence of BMs in recent years.[3,4] Concurrently, improvements in the sensitivity of medical imaging technologies have enhanced the detection of smaller and previously occult lesions.[5-7] The current standard of care for patients with a limited number of BMs is stereotactic radiosurgery (SRS), which offers excellent local control while sparing healthy tissue compared to whole-brain radiotherapy (WBRT).[8]

While highly effective, stereotactic radiosurgery (SRS) is associated with a significant challenge in patient follow-up.[9,10] The high focal radiation dose necessary for tumor ablation is associated with an elevated risk of radiation necrosis (RN), which is a delayed inflammatory tissue injury.[11,12] In parallel, the standard of care for patients treated with SRS includes frequent magnetic resonance imaging (MRI) monitoring to detect intracranial relapse. As a result of this treatment and monitoring strategy, clinicians are regularly presented with new or enlarging contrast-enhancing lesions on follow-up scans, which given their overlapping imaging characteristics on conventional MRI creates the challenge of differentiating tumor progression from RN.

Nowadays, histopathological analysis via surgical biopsy is considered the definitive gold standard for the diagnosis of the regrowth dynamics.[13] However, its clinical utility is severely constrained by numerous practical and biological limitations. Brain biopsy is an invasive and costly surgical procedure associated with inherent risks, and in certain anatomical locations, tissue accessibility may preclude its feasibility.[14]

In response to these limitations, advanced imaging modalities and computational approaches have been investigated to improve differentiation between tumor progression and RN. Perfusion-weighted MRI, MR spectroscopy, and amino acid positron emission tomography (PET) have shown potential in highlighting physiological or metabolic differences between necrotic and viable tumor tissue.[15,16] However, these techniques are not always accessible in routine clinical workflows and require specialized acquisition protocols.

The limitations of hypothesis-driven physiological imaging have led to a shift toward more agnostic, data-driven computational methods. This evolution began with radiomics-based analyses, which extract quantitative features from imaging data. These analyses have been applied to this problem with some success. Yet, these models often are prone to limited generalizability, as they are based on low-level imaging features, typically trained on small, single-center cohorts and rarely undergo independent multi-center validation.[17,18]

More recently, deep learning methods have shown promising performance improvements over hand-crafted radiomics. For instance, Ressa et al. recently demonstrated high diagnostic accuracy using a supervised CNN on a cohort of biopsy-proven lesions.[21] However, such fully supervised approaches are typically dependent on curated datasets. In neuro-oncology, where biopsy-confirmed labels are scarce and heterogeneous, this reliance on annotated data limits their generalizability and broader clinical adoption.[20]

Parallel to these efforts, the field of medical image analysis has witnessed transformative progress through self-supervised learning (SSL).[22] Unlike traditional supervised approaches, SSL leverages vast amounts of unlabeled imaging data by designing pretext tasks that encourage models to learn generalizable and anatomically meaningful representations. When subsequently fine-tuned on smaller labeled cohorts, SSL models often demonstrate superior performance and improved transferability across institutions.[23,24] This paradigm is particularly well suited to the problem of RN prediction, where labeled examples are rare but large collections of routine MRI scans are available.

Taken together, these developments highlight the need for clinically relevant, generalizable models that can distinguish radiation necrosis from tumor progression using routine post-contrast T1-weighted MRI. Beyond predictive accuracy, clinical adoption also requires interpretability. In this work, we address both by leveraging self-supervised representation learning for RN prediction and employing attention-based visualization to demonstrate that the model focuses on relevant brain regions, which enhances its transparency and potential clinical utility.

## Material and Methods

### Datasets and Population

#### Self-supervised Training

To develop a robust and generalizable feature representation, we carried out SSL on a large and heterogeneous collection of post-contrast T1-weighted (T1CE) brain MRIs from multiple sources. The pre-training cohort comprised 5 public datasets, namely Stanford BrainMetShare (n = 101),[25] NYUmets (n = 164),[26] The University of California San Francisco Brain Metastases Stereotactic Radiosurgery (UCSF-BMSR) (n = 323),[27] BraTS-METS 2023 (n = 238), Yale-Brain-Mets-Longitudinal (n = 200),[28] and an in-house dataset acquired at University Hospital Erlangen (n = 735), previously described.[29,30] From this corpus, a total of 10,167 lesion-centered sub-volume cubes (64×64×64 mm³) was extracted for pre-training, using available tumor segmentations to determine lesion centers. An institutional review board vote was available for all datasets. Moreover, approval was obtained from the ethics committee at Erlangen University Hospital to conduct this study.

#### Radiation Necrosis Classification

For fine-tuning and testing the self-supervised model, we utilized the MOLAB brain metastasis dataset ($n_{patients}$ = 75),[31] encompassing 260 brain metastases found in adult patients. The data was collected between 2005, and 2021, and had to include, at minimum, a high-resolution post-contrast T1-weighted imaging sequence. Ground-truth labels for this cohort were established through a combination of methods, including histopathology, metabolic imaging (PET), or definitive longitudinal MRI assessment (personal communication with the corresponding authors). To ensure data quality and consistency, we applied the following inclusion criteria: (i) the presence of at least one scan in which a metastasis exhibited ≥ 20% increase in longest diameter relative to nadir, with an absolute increase of at least 5 mm, as per RANO-BM assessment guidelines[32]; (ii) absence of neighboring metastases with conflicting progression labels within a 64×64×64 mm region centered on the lesion; and (iii) no severe motion artifacts. After applying these criteria, 109 metastases remained eligible for downstream stratified 5-fold cross-validation (80%) and same-center testing (20%, hold-out test set). In addition to imaging data, relevant clinical and treatment-related variables were collected for all patients, as described in the Feature Extraction and Selection section below.

For multicenter evaluation, a second independent dataset with 28 metastases from University Hospital Erlangen (UKER) was utilized. This dataset included n = 23 retrospective cases of progression and radiation necrosis as well as n = 5 enlarging brain metastases from the prospective randomized FSRT-Trial.[33] Ground-truth labels were determined using histopathological confirmation (biopsy or surgical resection, $n_{mets}$ = 9), longitudinal MRI assessment demonstrating consistent progression patterns or spontaneous resolution ($n_{mets}$ = 16), or metabolic characterization with amino acid PET imaging ($n_{mets}$ = 3). The aforementioned inclusion criteria were applied, and the institutional review board approval process was consistent with the other cohorts.

## Data Preprocessing

The T1CE images were reoriented to the LPS coordinate system, corrected for bias field inhomogeneities, and subsequently registered and resampled to an isotropic resolution of 1 mm³ using the SRI24 anatomical atlas as the reference standard.[34] Then, skull stripping was conducted using HD-BET brain extraction tool.[35] Finally, for ViT model input, z-score normalization was performed. The T1CE images and their corresponding binary lesion segmentation masks were both cropped to a 64x64x64 mm sub-volume centered on the lesion and concatenated to form a two-channel input tensor.

For the structured tabular features, the continuous values are normalized to have a mean of 0 and standard deviation of 1 based on the statistics of the training cohort. In contrast, categorical data were mapped to one-hot encoded vectors.

## Feature Extraction and Selection

In addition to the extracted sub-volumes, radiomic features were derived from the segmented tumors using the PyRadiomics package (version 2.2.0).[36] In addition to imaging-based features, we also incorporated clinical and treatment-related variables, including age, sex, primary tumor, systemic treatment, and recurrence time.

All variables underwent the same multi-step feature selection pipeline. Features with low variance (var < 0.01) across the cohort were first discarded. Highly correlated variables (Pearson's |r| > 0.9) were then removed to minimize redundancy. The remaining features were standardized (z-score normalization) and ranked for classification relevance using LASSO feature selection with 5-fold cross-validation, which retained 29 predictors from the original 1166 extracted features.

## Model Architecture and Training Process

To overcome the challenges of limited labeled data and the heterogeneity of MR Images, we employ a two-phase learning strategy as illustrated in Figure 1. The first phase involves the self-supervised pre-training of a Vision Transformer on a large corpus of unlabeled T1CE sub-volumes. Each sample comprises a T1CE image and its associated binary segmentation mask concatenated along the channel dimension, resulting in a two-channel input representation. Using context restoration and contrastive learning as pretext objectives, this initial stage utilizes self-attention to generate generalizable and discriminative feature representations.[24,37] A comprehensive description of the training methodology is provided in the Supplementary Materials. In the second phase, this pre-trained encoder is then fine-tuned, in a stratified 5-fold cross-validation setup, using supervised learning for distinguishing radiation necrosis from tumor progression. The model is then evaluated on the held-out test set from the MOLAB dataset, as well as on the second-center UKER dataset for multi-center evaluation.

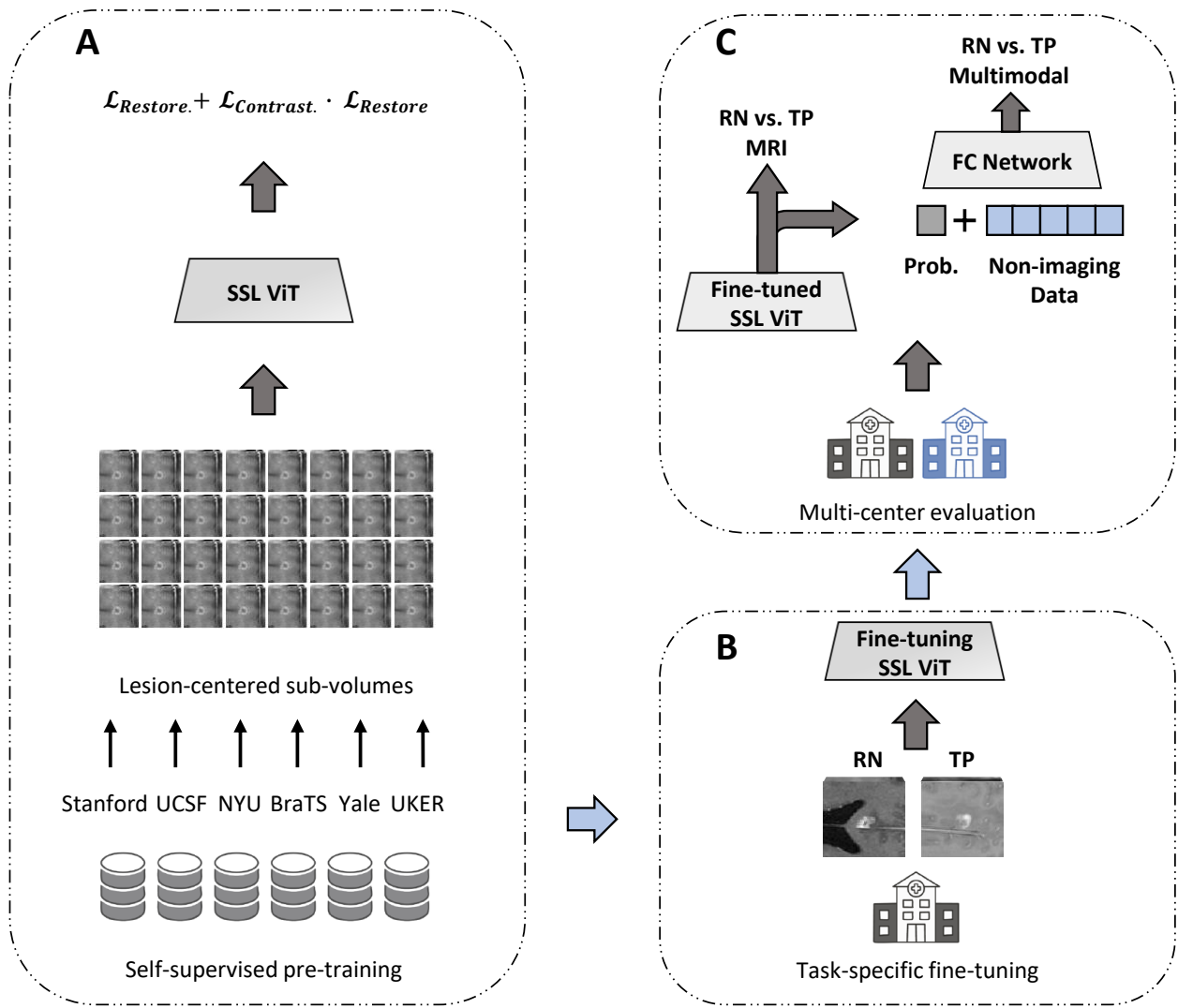

*Figure 1 Workflow of the proposed two-phase deep learning strategy. The process begins with (A) self-supervised pre-training, where the model learns robust, general features from a large corpus of unlabeled brain metastases datasets. Subsequently, the pre-trained model is adapted via (B) Fine-tuning on the data-scarce target task: differentiation of radiation necrosis from tumor progression. C) Finally, the model is evaluated in a multi-center evaluation using a same-center test set as well as second-center test data.*

## Multimodal Integration

To effectively integrate the rich, high-dimensional information from the imaging sub-volumes with the structured patient data, we employed a model stacking approach. This technique uses the out-of-fold output from our fine-tuned ViT as a single, high-level feature. This imaging-derived probability feature was then concatenated with the pre-extracted clinical and radiomic features. This final, combined feature set was used to train a multi-layered fully connected classifier to produce the definitive classification of radiation necrosis or tumor progression.

## Ablation Study

To validate the impact of the proposed approach, we performed two ablation studies. First, we quantified the performance gain from SSL pre-training by training a fully-supervised standard 3D ViT from scratch using only the labeled MOLAB training data. Second, we assessed the value of including the tumor segmentation into the model input. We compared our main two-channel (T1CE + Segmentation) SSL ViT against an "SSL ViT (T1CE-only)" model. This ablation model used the same SSL pre-training but was fine-tuned using only the T1CE sub-volume as input.

## Model Interpretation and Explainability

To allow for interpretability, attention weight maps from the Vision Transformer were visualized to highlight image regions most influential in classification. This allowed to qualitatively assess whether the model focused on clinically relevant structures and gain mechanistic insights.

## Statistical Analysis

Clinical characteristics were compared using Fisher's exact tests for categorical variables, whereas continuous variables were assessed for normal distribution using the Shapiro-Wilk test. Comparisons between non-normally distributed groups were performed using a Mann-Whitney U test, while a paired t-test was used to evaluate differences in model performance metrics. For all statistical tests, a p-value < 0.05 was considered statistically significant.

# Results

## Cohorts Characteristics

The baseline patient and treatment characteristics for both cohorts are summarized in Table 1. A total of 109 lesions from the MOLAB dataset and 28 lesions from the UKER dataset were analyzed. No significant differences were observed between the two datasets in the mean age of patients (56.5 years for MOLAB vs. 56.5 years for UKER, p = 0.794) or in the proportion of lesions pathologically confirmed as RN (33.9% vs. 39.3%, p = 0.597). However, the cohorts differed significantly in some characteristics. The distribution of primary cancer types was highly dissimilar (overall p < 0.0001; the MOLAB dataset was predominantly composed of NSCLC (49.5%) and breast cancer (33.0%), whereas the UKER dataset was dominated by melanoma (53.5%). Furthermore, the MOLAB dataset included a significantly smaller proportion of male patients (35.7% vs. 64.3%, p = 0.010). Prior WBRT before SRS was also significantly more common in the MOLAB cohort compared to the UKER cohort (49.5% vs. 17.9%, p = 0.002). Further details regarding the baseline characteristics with respect to pathological classification are available in the Supplementary Materials (Supplementary Table S1).

*Table 1 Summary of patient and treatment characteristics for the two cohorts:*

| Parameter | MOLAB Dataset (n = 109) | UKER Dataset (n = 28) | P for difference |
|---|---|---|---|
| Pathology, n (%) | | | 0.597 |
|   Radiation Necrosis (RN) | 37 (33.9%) | 11 (39.3%) | |
|   Progression | 72 (66.1%) | 17 (60.7%) | |
| Sex, n (%) | | | 0.010 |
|   Male | 40 (36.7%) | 18 (64.3%) | |
|   Female | 69 (63.3%) | 10 (35.7%) | |
| Age, years | | | 0.794 |
|   Mean (range) | 56.5 (27 – 77) | 56.5 (35 – 78) | |
| WBRT before SRS, n (%) | | | 0.002 |
|   Yes | 54 (49.5%) | 5 (17.9%) | |
|   No | 55 (50.5%) | 23 (82.1%) | |
| Primary Cancer, n (%) | | | < 0.0001 |
|   NSCLC | 54 (49.5%) | 3 (10.7%) | |
|   Breast | 36 (33.0%) | 5 (17.9%) | |
|   Melanoma | 8 (7.3%) | 15 (53.5%) | |
|   SCLC | 7 (6.4%) | 0 (0%) | |
|   Ovary | 2 (1.9%) | 0 (0%) | |
|   Other* | 2 (1.9%) | 5 (17.9%) | |

\* Primary cancer types that appeared only once (uterus and bones in the MOLAB dataset) or were unspecified (in the UKER dataset).

## Imaging Model Comparison

Our evaluation demonstrates the performance of the pre-trained self-supervised Vision Transformer (SSL ViT) compared to other imaging-based models for differentiating radiation necrosis from tumor progression in brain metastases. On the MOLAB test set, the SSL ViT model achieved an Area Under the Curve (AUC) of 0.916. This was a notable improvement over the traditional Random Forest-based radiomics model (AUC 0.807; p=0.005), and the ResNet50 model (AUC 0.637; p=0.003), which followed the implementation of Ressa and coauthors.[21] The SSL ViT also obtained the highest accuracy at 0.909 and a particularly high sensitivity of 0.969, while maintaining a specificity of 0.866. The SSL ViT model's superior performance was confirmed on the UKER test set. It achieved an AUC of 0.764, outperforming the radiomics model (AUC 0.691; p=0.014) as well as ResNet50 (AUC 0.525; p<0.001). The model's accuracy on this UKER cohort was 0.765, with a sensitivity of 0.744 and a specificity of 0.770. These results are summarized in Table 2 and ROC curves are illustrated in Figure 2(A-B). For the subset of n = 9 histopathologically confirmed cases in the second-center test set, the SSL ViT model achieved an accuracy of 0.889 (8/9 correctly classified lesions).

## Multimodal Integration Performance

Integrating the SSL ViT with clinical and radiomic features further enhanced the model's predictive capabilities, resulting in the highest performance across all experiments. On the MOLAB test set, the full multimodal model (SSL ViT + Clinical + Radiomics) reached an AUC of 0.947 and an accuracy of 0.927. This surpassed the performance of the SSL ViT alone (AUC 0.916; p=0.073), the Radiomics + Clinical model (AUC 0.886), and the model using only clinical data (AUC 0.711). This advantage was also evident on the UKER test set, where the multimodal model achieved an AUC of 0.821 and an accuracy of 0.782. This again represented the best performance, outperforming the standalone SSL ViT (AUC 0.764; p=0.001) and the combined Radiomics + Clinical model (AUC 0.706). These results are summarized in Table 3 and ROC curves are illustrated in Figure 2 (C-D). On the histopathologically confirmed UKER subset, the multimodal model achieved an accuracy of 0.889 (8/9 samples), matching the performance of the imaging-only model.

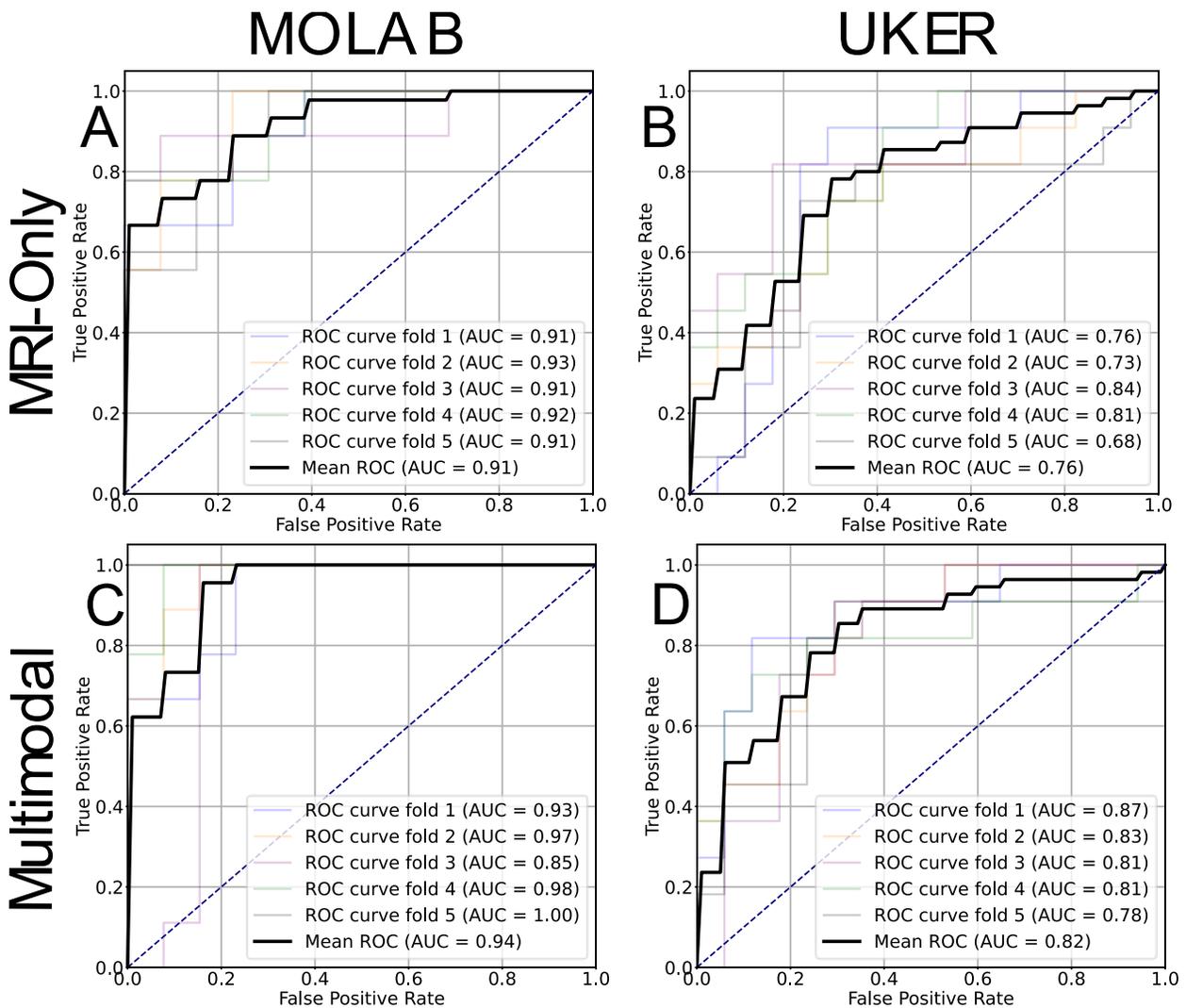

*Figure 2 ROC curve analysis comparing model performance across MOLAB and UKER test sets. The top row displays results for the Imaging-only model, while the bottom row shows the multimodal model. Both models were evaluated on the MOLAB test sets A and C and the UKER test sets B and D.*

Table 2 Performance metrics for the monomodal models on the MOLAB and UKER test sets utilizing imaging / imaging-derived data from post-contrast T1-weighted MRI only.

|  |  | Radiomics | ResNet50[21] | SSL ViT |
|---|---|---|---|---|
| MOLAB Test Set (Same center) | AUC | 0.807 ± 0.046 | 0.637 ± 0.107 | **0.916 ± 0.008** |
|  | Accuracy | 0.768 ± 0.054 | 0.690 ± 0.066 | **0.909 ± 0.022** |
|  | Sensitivity | 0.657 ± 0.145 | 0.466 ± 0.257 | **0.969 ± 0.037** |
|  | Specificity | 0.883 ± 0.066 | 0.846 ± 0.137 | 0.866 ± 0.044 |
| UKER Test Set (Second center) |  | Radiomics | ResNet50 | SSL ViT |
|  | AUC | 0.691 ± 0.046 | 0.525 ± 0.042 | **0.764 ± 0.056** |
|  | Accuracy | 0.627 ± 0.041 | 0.593 ± 0.014 | **0.765 ± 0.024** |
|  | Sensitivity | 0.350 ± 0.050 | 0.720 ± 0.257 | **0.744 ± 0.012** |
|  | Specificity | **0.850 ± 0.044** | 0.488 ± 0.206 | 0.770 ± 0.004 |

Table 3 Performance metrics for the multimodal imaging, clinical and radiomics-based models on the MOLAB and UKER test sets compared to monomodal models using imaging and clinical data alone, respectively.

|  |  | Imaging | Non-imaging | Multimodal | |
|---|---|---|---|---|---|
|  |  | SSL ViT | Clinical | Radiomics + Clinical | SSL ViT + Clinical + Radiomics |
| MOLAB Test Set (Same center) | AUC | 0.916 ± 0.008 | 0.711 ± 0.052 | 0.886 ± 0.068 | **0.947 ± 0.019** |
|  | Accuracy | 0.909 ± 0.022 | 0.706 ± 0.069 | 0.842 ± 0.110 | **0.927 ± 0.036** |
|  | Sensitivity | **0.969 ± 0.037** | 0.514 ± 0.429 | 0.629 ± 0.265 | 0.889 ± 0.000 |
|  | Specificity | 0.866 ± 0.044 | 0.700 ± 0.281 | **0.967 ± 0.040** | 0.954 ± 0.062 |
|  |  | SSL ViT | Clinical | Radiomics + Clinical | SSL ViT + Clinical + Radiomics |
| UKER Test Set (Second center) | AUC | 0.764 ± 0.056 | 0.636 ± 0.019 | 0.706 ± 0.054 | **0.821 ± 0.029** |
|  | Accuracy | 0.765 ± 0.024 | 0.624 ± 0.014 | 0.607 ± 0.117 | **0.782 ± 0.049** |
|  | Sensitivity | 0.744 ± 0.012 | 0.760 ± 0.053 | 0.763 ± 0.178 | **0.799 ± 0.015** |
|  | Specificity | **0.770 ± 0.004** | 0.511 ± 0.022 | 0.506 ± 0.287 | 0.769 ± 0.084 |

## Ablation Experiments and Model Explainability

Subsequently, we conducted ablation experiments to evaluate the individual contributions of self-supervised pre-training and segmentation-guided learning, respectively. The fully-supervised ViT trained from scratch without pre-training performed poorly, achieving AUCs of only 0.624 on the MOLAB test set (p = 0.001) and 0.496 on the UKER test set (p = 0.008). These low scores emphasize the value of large-scale pre-training in this low-data regime. We also compared our main two-channel SSL ViT (T1CE + lesion segmentation) to an ablated model fine-tuned solely on T1CE images. The single-channel SSL ViT achieved substantially lower AUC scores (MOLAB: 0.895 vs. 0.916; UKER: 0.717 vs. 0.764), demonstrating the added value of incorporating segmentation masks into the model input.

The importance of the segmentation input is further evident in the model's attention maps. In cases with relatively large lesions (Figures 3A-B), both models - with and without segmentation input - correctly attended to lesion-containing regions. However, in cases with small lesions adjacent to hyperintense anatomical structures such as the dura and venous sinuses (Figures 3C-D), critical differences emerged. The model trained on T1CE images alone misidentified these hyperintense normal structures as lesions. In contrast, the model incorporating tumor segmentations focused on the true tumor regions, avoiding confounding anatomical features.

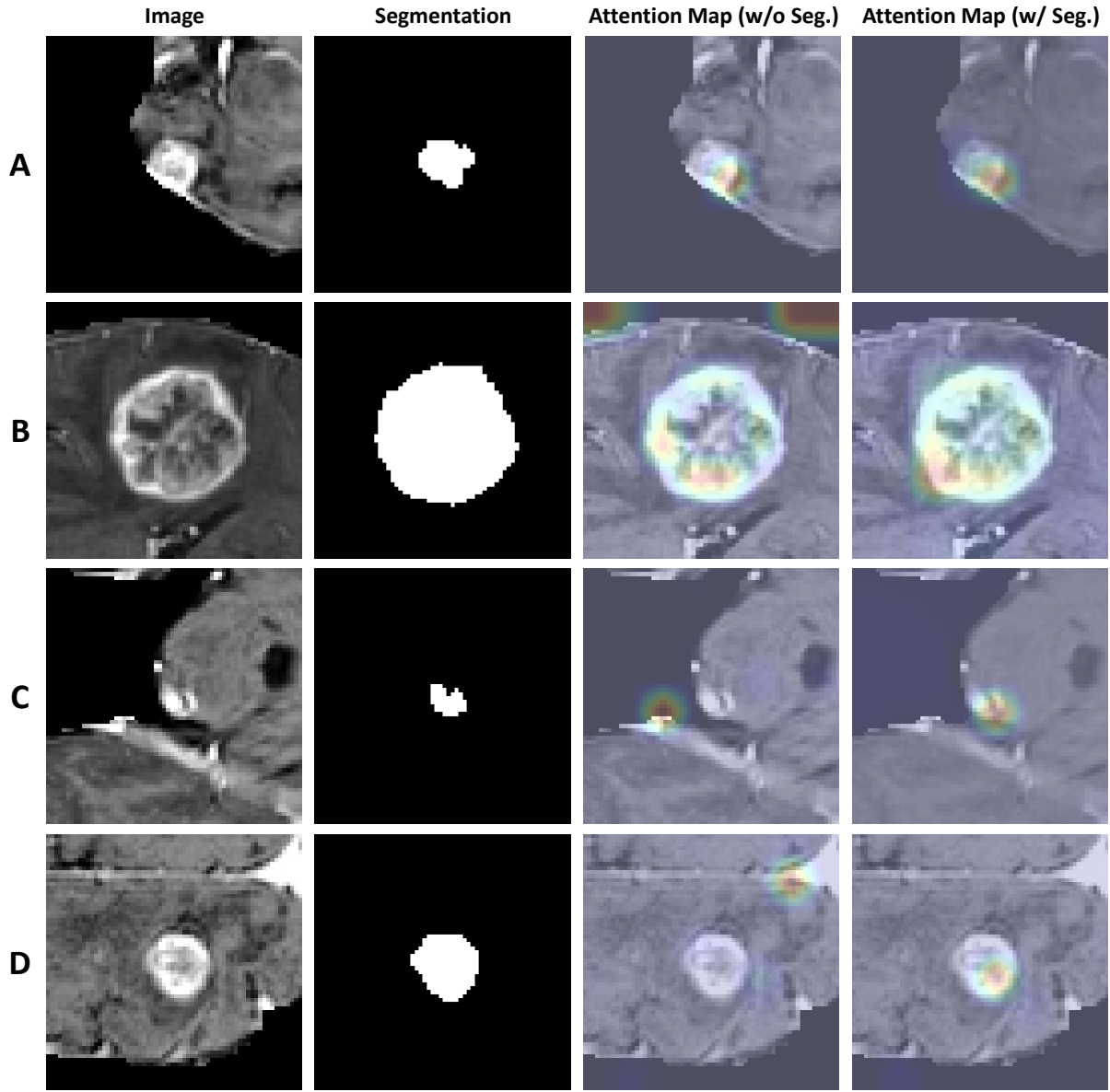

*Figure 3 Attention map visualizations for model explainability on representative test cases. The figure compares attention patterns between the model trained on T1CE images alone and the model trained on T1CE images with corresponding tumor segmentations. (A-B) Cases with large, well-defined lesions (A, tumor progression and B, radiation necrosis). Both models correctly attend to the lesion regions. (C-D) Cases with smaller lesions adjacent to hyperintense anatomical structures (C, tumor progression and D, radiation necrosis). Only the segmentation-guided model shows correct attention, highlighting how the segmentation input enables the model to distinguish the relevant lesion from confounding anatomical structures.*

## Discussion

The differentiation of radiation necrosis from tumor progression remains a major challenge in the post-treatment management of brain metastases. This distinction is critical as it may lead to very different treatment paths, from observation and medical management for RN to re-irradiation for true progression. The current gold standard, histopathological analysis by stereotactic biopsy or lesion resection, is an invasive surgical procedure associated with inherent risks and is not always feasible.[38] Consequently, the development of a robust, non-invasive biomarker to accurately identify these uncertain lesions is of high clinical importance.

In this work, we demonstrate that a self-supervised ViT, pre-trained on a large, diverse corpus of unlabeled T1CE MR images can accurately distinguish radiation necrosis from tumor progression in patients with brain metastases treated with stereotactic radiosurgery. Compared to conventional radiomics and fully supervised deep learning approaches, the SSL ViT achieved notably improved predictive performance on both MOLAB and UKER test cohorts. This improvement highlights the advantages of self-supervised representation learning in contexts where labeled data is limited and heterogeneous, conditions that are typical of post-radiosurgery brain metastasis imaging. Furthermore, when imaging-derived deep features were integrated with complementary clinical and radiomic variables within a multimodal framework, the model achieved the highest overall predictive accuracy. This result demonstrates the synergistic value of combining quantitative imaging and clinical information for robust post-treatment response assessment.

A detailed look at the results reveals the critical advantage of our self-supervised pre-training strategy. The SSL ViT model's strong performance stands in contrast to results from the models trained from scratch. Both the fully-supervised 3D ResNet50 (AUC 0.637 MOLAB, 0.525 UKER) and the ViT (AUC 0.624 MOLAB, 0.496 UKER) were not as effective as the proposed pre-trained SSL ViT in learning discriminative features from the data, performing worse than even the traditional radiomics model. This confirms our hypothesis that in this heterogeneous and low-data regime where labeled examples are rare, standard supervised training is insufficient, and large-scale self-supervised pre-training is a critical strategy to enable deep learning models to learn robust, generalizable features.[39-41]

Our findings are especially clear when compared to the performance of traditional radiomics, which has been the focus of many previous efforts.[42-44] Our results agree with what is often observed: radiomics-based models have limited generalizability in multi-center setups.[45] While our optimized radiomics model achieved favorable performance on the MOLAB test set (AUC 0.807), its performance saw a clear drop on the UKER cohort (AUC 0.691). This drop is a known problem with handcrafted features, which can be sensitive to variations in acquisition parameters and patient cohorts.[46] In contrast, the SSL ViT was more reliable on the UKER cohort (AUC 0.764). This result demonstrates its ability to learn higher-level and transferable imaging features directly from the data.[47]

While the SSL ViT provides a strong imaging-based foundation, our results also highlight the value of multimodal integration. The highest overall performance (AUC 0.947 MOLAB, 0.821 UKER) was achieved by combining the deep learning features with select clinical and radiomic data. This combination suggests that these data sources provide complementary information, and their integration is key to building the most robust predictive tool. A notable advantage of this multimodal framework is that it is based on easily accessible parameters. It relies only on conventional T1CE MRI, the standard sequence for BM imaging, and routine clinical variables, without the need for specialized or advanced imaging techniques like perfusion-weighted MRI, MR spectroscopy, or amino acid PET. By relying solely on standard-of-care T1CE

MRI and clinical data, this framework lowers barriers to clinical adoption and enables easier validation across diverse institutional settings.

Beyond predictive accuracy, the "black box" nature of deep learning remains a major barrier to clinical adoption. The use of attention-based visualizations, as shown in Figure 3, provides an important level of understanding. These maps confirm that the SSL ViT learns to focus on clinically relevant structures within and around the lesion, rather than confounding anatomy. They also suggest that the segmentation-guided model attends better to smaller lesions adjacent to hyperintense meningeal and vascular structures than the model trained with MRI sub-volumes alone. Interestingly, these attention maps also provide mechanistic insights into which lesion sub-compartments drive the model's predictions, patterns that can be correlated with existing clinical knowledge while potentially helping to uncover new discriminative features. The attention maps show that the model consistently concentrates more on the enhancing rim and the immediate perilesional interface rather than the necrotic center or distant parenchyma. This pattern is consistent with reported morphologic differences between radiation necrosis and tumor progression. Previous studies have shown that radiation necrosis is more frequently associated with a discontinuous enhancing rim ("open ring" pattern), while tumor progression tends to demonstrate thick nodular or solid enhancement patterns.[48] Moreover, radiation necrosis has been linked to a "soap-bubble"-like interior (Figure 3B).[49,50] Taken together, these findings indicate that attention-based visualization not only enhances interpretability but also reinforces confidence in the model's reasoning by allowing clinicians to verify that its focus aligns with known pathophysiologic features.

Although the SSL ViT demonstrated significantly improved generalizability compared to conventional methods, the observed performance attenuation between the internal and external validation cohorts (AUC 0.916 vs. 0.764) may be attributed to several confounding factors. First, the reference standards utilized to establish ground truth were heterogeneous. They relied on a composite of histopathology, PET imaging, and longitudinal MRI assessment rather than uniform biopsy confirmation. This methodological variance introduces a degree of label uncertainty that may impact evaluation metrics across different centers. Second, the cohorts exhibited significant divergence in baseline clinical statistics. Specifically, the internal cohort was predominantly composed of NSCLC and breast cancer patients with a higher incidence of prior WBRT, whereas the external cohort was dominated by melanoma cases. Finally, despite standardized preprocessing, the inherent variability in MRI acquisition protocols and scanner characteristics across institutions introduces domain shifts in signal intensity distributions that remain a challenge in medical image analysis.

This study is not without limitations. While the training and same-center hold-out test set was based on biopsy-proven labels, the ground-truth for a subset of the second-center test cohort was based on longitudinal MRI assessment and FET-PET rather than histopathological confirmation. While the histopathologically confirmed subset was limited (n = 9), the self-supervised model achieved 88.9% accuracy against this gold standard reference. Furthermore, while we validated on an external center dataset, future work should focus on prospective validation on data from additional centers. Further investigations could also look at adding other commonly-available routine MRI sequences (e.g., T2-FLAIR) into the self-supervised and multimodal framework to see if they capture additional complementary information.

In conclusion, we describe a novel two-phase multimodal deep learning strategy, centered on a self-supervised Vision Transformer, that accurately differentiates radiation necrosis from tumor progression. By leveraging large-scale unlabeled data for pre-training, this approach can overcome data scarcity in brain metastases and provides a promising, interpretable framework for a pressing challenge in neuro-oncology.

# Acknowledgments

The present work was performed in partial fulfillment of the requirements for obtaining the degree „Dr. rer. biol. hum." at the Friedrich-Alexander-Universität Erlangen-Nürnberg (FAU).

# Authorship Statement

The study was designed and conceived by F.P., A.S. and A.G. A.G. wrote the program code, conducted the experimental work. A.G., A.S., and F.P. wrote the original draft. L.S., A.D., M.S.M., P.S., I.S., J.S., K.B., Y.H., B.F., O.S., D.D., R.S., D.H., P.S., J.S., S.S., A.M., D.H.H., U.S.G, R.F., C.B., and S.C. reviewed and edited the script. All authors have read and approved the script.

## Funding

Open Access funding enabled and organized by Projekt DEAL.

## Conflict of interest statement

The authors declare no conflict of interest in this work.